\documentclass{article} 
\usepackage{iclr2027_conference,times}

\usepackage{amsmath,amsfonts,bm}

\def\eqref#1{equation~\ref{#1}}

\def\1{\bm{1}}

\DeclareMathAlphabet{\mathsfit}{\encodingdefault}{\sfdefault}{m}{sl}
\SetMathAlphabet{\mathsfit}{bold}{\encodingdefault}{\sfdefault}{bx}{n}

\usepackage[table,dvipsnames]{xcolor}
\definecolor{mydarkblue}{rgb}{0,0.08,0.45}
\usepackage{url}
\usepackage{booktabs}
\usepackage{lineno}
\usepackage{amssymb}
\usepackage{subcaption}
\usepackage{graphicx}
\usepackage{amsmath}
\usepackage{wrapfig}
\usepackage{multirow}
\usepackage{enumitem}
\usepackage{siunitx}

\usepackage[colorlinks=true, linkcolor=mydarkblue, citecolor=mydarkblue,
            urlcolor=mydarkblue, filecolor=mydarkblue]{hyperref}
\usepackage{cleveref}

\usepackage{todonotes}
\title{Shifting Mechanisms: How Positional \\Encoding Choice Shapes In-Context Retrieval}

\author{Eric Enouen \\
  Cornell University \\
  \texttt{enouen@cs.cornell.edu} \\
  \And
  Sainyam Galhotra \\
  Cornell University \\
  \texttt{sg@cs.cornell.edu} \\
}

\usepackage{fancyhdr}
\fancypagestyle{arxiv}{%
  \fancyhf{}                          
  \fancyfoot[C]{\thepage}             
  \renewcommand{\headrulewidth}{0.4pt} 
}

\iclrfinalcopy 
\begin{document}
\newcommand{\Comment}[1]{}

\newcommand{\bhl}[1]{{\color{purple}#1}}
\newcommand{\dep}[1]{{\color{red}#1}}
\definecolor{ultrapink}{rgb}{1.0, 0.44, 1.0}
\definecolor{neongreen}{rgb}{0.6, 0.9, 0}
\definecolor{ericgreen}{rgb}{0, .65, 0}

\newcommand{\todoc}[2]{{\textcolor{#1}{\textbf{#2}}}}
\newcommand{\todored}[1]{{\todoc{red}{\textbf{[[#1]]}}}}
\newcommand{\todogreen}[1]{\todoc{green}{\textbf{[[#1]]}}}
\newcommand{\todoblue}[1]{\todoc{blue}{\textbf{[[#1]]}}}
\newcommand{\todoorange}[1]{\todoc{orange}{\textbf{[[#1]]}}}
\newcommand{\todobrown}[1]{\todoc{brown}{\textbf{[[#1]]}}}
\newcommand{\todogray}[1]{\todoc{gray}{\textbf{[[#1]]}}}
\newcommand{\todopink}[1]{\todoc{purple}{\textbf{[[#1]]}}}
\newcommand{\todopurple}[1]{\todoc{ultrapink}{\textbf{[[#1]]}}}
\newcommand{\todoneongreen}[1]{\todoc{neongreen}{\textbf{[[#1]]}}}
\newcommand{\todoericgreen}[1]{\todoc{ericgreen}{\textbf{[#1]}}}

\definecolor{light-gray}{gray}{0.7}
\newcommand{\hilight}[1]{\colorbox{light-gray}{#1}}


\newcommand{\sg}[1]{\textcolor{blue}{{[SG: #1]}}}
\newcommand{\eric}[1]{\todoericgreen{Eric: #1}}

\newcommand{\code}[1]{\begin{small}\texttt{#1}\end{small}\xspace}
\newcommand{\q}[1]{``#1''\xspace}

\newcommand{\ck}[1]{\todo{check->}#1}
\renewcommand{\ck}[1]{#1}

\newcommand{\dlt}[1]{{\scriptsize\color{red!70!black}\num[retain-explicit-plus]{#1}}}
\newcommand{\dgt}[1]{{\scriptsize\color{green!50!black}\num[retain-explicit-plus]{#1}}}
\newcommand{\dltsmall}[1]{\textcolor{red!40}{\scriptsize #1}}
\newcommand{\sd}[1]{\,{\scriptsize(#1)}}
\newcommand{\rpos}{\ensuremath{r_{\mathrm{pos}}}}
\definecolor{ropeblue}{HTML}{1F5DB1}    
\definecolor{hybridsalmon}{HTML}{E8646D} 

\definecolor{darkpurple}{HTML}{4B0082} 
\definecolor{darkblue}{HTML}{1155CC}
\definecolor{darkgreen}{HTML}{38761D}

\pagestyle{arxiv}
\maketitle
\thispagestyle{arxiv}

\begin{abstract}
Language models increasingly use architectures that vary attention span and positional encoding across layers, such as applying RoPE with sliding-window attention and NoPE with global attention (SWA NoPE). However, how these choices shape in-context retrieval remains unclear. 
To study this question, we take a mechanistic view, tracing how positional encoding (PE) choice shapes the internal mechanisms models use for in-context retrieval.
Across 22 open-weight models spanning eight families, we find that standard RoPE models rely primarily on positional retrieval, while PE hybrids shift toward semantic retrieval. 
We further show on a controlled pre-training ablation that confining positional encoding to local layers produces this semantic shift, degrading representations of positional information.
Finally, we show that the reported long-context gains of PE hybrids mask a retrieval trade-off: SWA NoPE improves over RoPE on multiple-target retrieval and QA, but degrades when distinguishing competing keys. We show that these behavioral differences better track the mechanism shift from positional toward semantic mechanisms than a uniform improvement in long-context retrieval.
\end{abstract}

\section{Introduction}
\label{sec:introduction}
Language model architectures have increasingly moved away from standard RoPE \citep{rope} toward hybrid designs that vary attention span and positional encoding across layers, architectural choices we jointly refer to as positional encoding (PE) choice. Prior work has shown that PE hybrids can improve long-context retrieval performance: SWA NoPE \citep{puvvada2025swan, yang2025hybrid}, which combines local sliding-window attention (SWA) layers using RoPE with global attention layers without PE (NoPE), outperforms RoPE on several long-context benchmarks. Yet the underlying explanation for how PE choice shapes long-context retrieval ability remains unclear.

Recent interpretability work provides a way to investigate this question, identifying three core mechanisms that language models use for in-context retrieval: positional, based on an entity's location in context, and lexical and reflexive, based on an entity's semantic content \citep{gur2025mixing}. Using this mechanistic analysis as an intermediate lens between model architecture and retrieval behavior, we ask whether PE choice changes how models retrieve information and whether these underlying mechanism differences can explain long-context retrieval performance.

We first find that PE hybrids have a consistently more semantic mechanism allocation. Across 22 open-weight models spanning eight families, RoPE models rely primarily on the positional mechanism, while PE hybrids rely primarily on semantic (lexical and reflexive) mechanisms. To study why this shift occurs, we analyze a controlled set of pre-training ablations of PE choice (from RoPE to SWA RoPE to SWA NoPE), finding that SWA alone does not produce the shift, while additionally removing RoPE from global layers does. We find confining positional encoding to local layers degrades representations of positional information, matching the observed semantic shift.

Finally, we show how our mechanistic analysis explains the impact of PE choice on long-context retrieval behavior. We find that SWA NoPE outperforms RoPE on multiple-target needle-in-a-haystack (NIAH) and QA tasks, where semantic mechanisms excel at finding relevant content, but degrades when trying to distinguish between semantically similar keys. These results demonstrate that retrieval strategies produce different strengths and failure modes at long context, suggesting that SWA NoPE's long-context gains better reflect a shift in retrieval strategy than a uniform improvement.

We summarize our contributions below:
\begin{itemize}[leftmargin=*, labelsep=4pt, itemindent=0pt, itemsep=2pt, parsep=0pt, topsep=0pt]
    \item \textbf{Positional Encoding Choice $\to$ Mechanism Allocation.} We show that retrieval strategies are architecture-dependent and replacing uniform RoPE with SWA NoPE systematically shifts mechanism allocation from positional toward semantic retrieval.
    \item \textbf{Why does this shift occur?} Linear probing suggests that confining positional encoding to local layers degrades representations of positional information, providing a potential explanation for the shift away from positional retrieval.
    \item \textbf{Mechanism Allocation $\to$ Long-Context Retrieval.} We show that the mechanism shift predicts long-context retrieval behavior. The shift toward semantic mechanisms improves retrieval of relevant content but increases sensitivity to semantic confusability, which we validate experimentally across long-context retrieval tasks.
\end{itemize}

\begin{figure}[t]
    \centering
    \includegraphics[width=\textwidth]{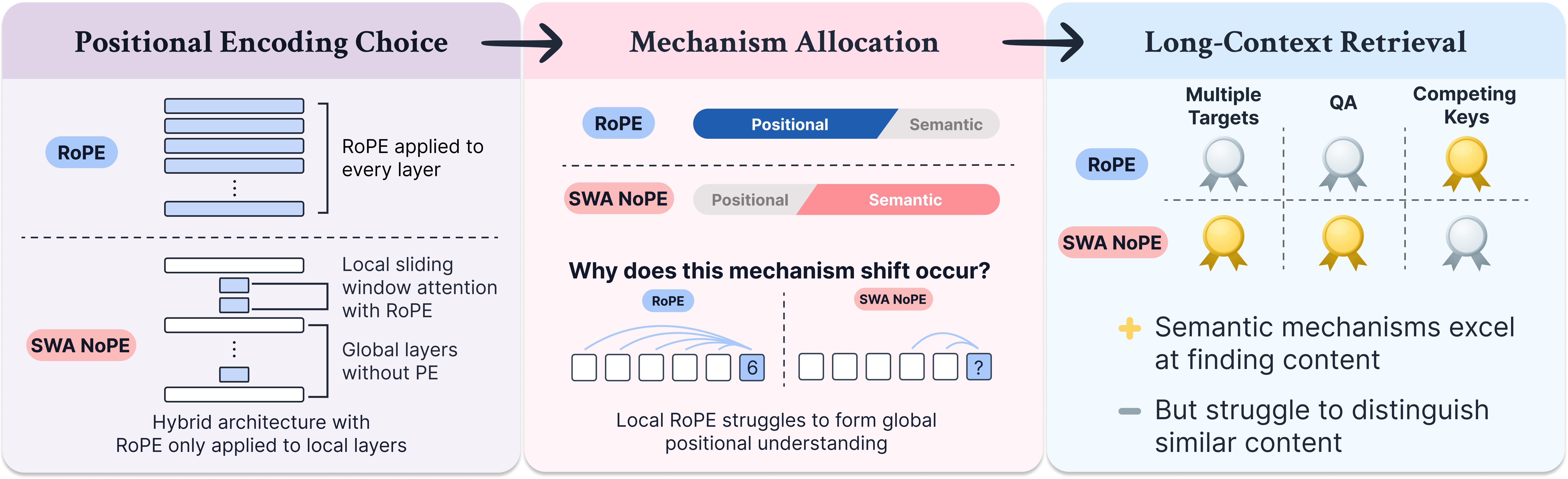}
    \caption{\textbf{Overview.} We show that positional encoding (PE) choice shapes mechanism allocation: the SWA NoPE architecture shifts toward semantic mechanisms compared to RoPE. We then show that this mechanism shift predicts long-context retrieval performance: SWA NoPE improves on multiple-target retrieval and QA tasks, but struggles to distinguish competing keys.}
    \label{fig:motivation}
\end{figure}

\section{Related Work}
\label{sec:related_work}
\textbf{Positional Encoding and Hybrid Architectures.}
RoPE \citep{rope} is widely used in modern LLMs, but can struggle to extrapolate beyond its training window \citep{liu2026rotary, du2025context, du2026rope}, motivating substantial work on how positional encoding can scale to longer contexts. One line of work extends RoPE through interpolation and frequency scaling \citep{blocntkaware, blocntkparts, emozillareddit, peng2024yarn}, while another modifies or removes explicit positional encoding altogether \citep{gelberg2025extending, barbero2024round, khan2026fractional, gopalakrishnan2025decoupling, movahedi2026selective}. Furthermore, transformers without explicit positional encodings can still achieve competitive length generalization in some settings \citep{haviv2022transformer, kazemnejad2023impact}.

Most relevant to our work are the hybrid SWA NoPE architectures proposed by \citet{yang2025hybrid, puvvada2025swan} that interleave local RoPE and global NoPE, improving aggregate long-context performance. Prior work has sought to explain these gains through properties of the architecture. \citet{qiao2026rethinking}, for example, argue that long-range retrieval is primarily carried by global attention layers, while local attention shapes how these retrieval capabilities emerge during training, motivating the use of NoPE in global layers. In contrast, we find that SWA NoPE does not uniformly improve long-context retrieval. Instead, its mechanism shift from positional toward semantic retrieval predicts distinct strengths and failure modes across retrieval settings.

\textbf{Mechanistic Analysis of In-Context Retrieval.}
Early mechanistic work on entity binding identified a positional retrieval mechanism \citep{feng2024language, prakash2024fine, prakash2025language}. More recently, \citet{gur2025mixing} showed that models also use lexical and reflexive mechanisms, which we refer to collectively as semantic mechanisms. Complementary work characterizes individual attention heads as positional or symbolic, finding theoretically and empirically that symbolic mechanisms generalize more robustly to longer sequences \citep{urrutia2025decoupling, urrutia2026positional}. While this work characterizes the mechanisms models use for retrieval, we study how architectural choices in positional encoding shape which of these mechanisms models learn to rely on.
\section{Mechanistic Lens}
\label{sec:mechanistic_lens}
In this section we explain the mechanistic lens we use throughout this work. Prior work \citep{gur2025mixing} found three core mechanisms that language models use to bind and retrieve entities in-context. We first describe the counterfactual approach used to disentangle these mechanisms, and then explain how we use it to measure mechanism allocation.\footnote{Code is available at {\href{https://github.com/ericenouen/shiftmech}{\textcolor{blue}{\texttt{https://github.com/ericenouen/shiftmech}}}}.}

\begin{figure}[t]
    \centering
    \includegraphics[width=\textwidth]{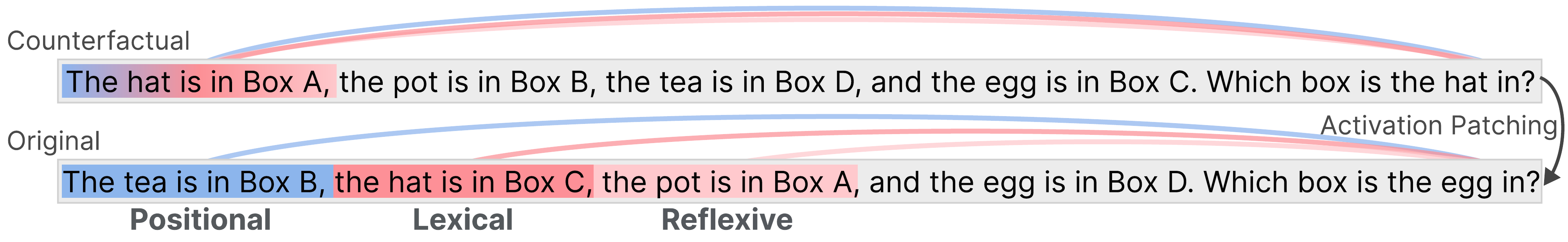}
    \caption{\textbf{Example of counterfactual activation patching.} The prompt pair is constructed so that each retrieval mechanism corresponds to a different answer after activation patching. The positional mechanism follows location and predicts `B', the lexical mechanism follows the `hat' key and predicts `C', and the reflexive mechanism follows the self-referential pointer and predicts `A'. If patching has no effect, the model retains the original prediction (`D').}
    \label{fig:background}
\end{figure}

\textbf{Counterfactual Patching.}
To understand how a model uses these mechanisms, \citet{gur2025mixing} use carefully paired original and counterfactual prompts and apply activation patching on the residual stream to identify which mechanism dominates a given prediction. We illustrate their counterfactual patching framework using the Boxes task in Figure~\ref{fig:background}, where objects are assigned to boxes and the model is queried for which box contains a particular object.

The prompts are constructed so that the patched prediction uniquely identifies the dominant mechanism. We define each mechanism below:
\begin{itemize}[leftmargin=*, labelsep=4pt, itemindent=0pt, itemsep=2pt, parsep=0pt, topsep=0pt]
    \item \textbf{Positional Mechanism (P).} The model stores the position of the queried item and later retrieves from that same position. In \Cref{fig:background}, the counterfactual answer `A' appears first, so the mechanism retrieves from the first position in the original prompt, returning `B'.
    \item \textbf{Lexical Mechanism (L).} The model stores the queried item and later retrieves using that item. In \Cref{fig:background}, the counterfactual answer stores the semantic key `hat', so the mechanism retrieves the box associated with `hat' in the original prompt, returning `C'.
    \item \textbf{Reflexive Mechanism (R).} The model stores a self-referential pointer to the answer and later retrieves that answer directly. In \Cref{fig:background}, the self-referential pointer directly stores `A', so the mechanism returns `A'.
\end{itemize}

In addition to these three mechanisms, patching can produce two other outcomes. We classify a sample as \textit{no effect} when the model retains the original prompt's prediction (`D' in \Cref{fig:background}), and as \textit{unknown} when the prediction does not correspond to any of the three mechanisms or the original prediction. With a larger number of entities (we study 20 entities throughout this work), there are many additional predictions that can therefore fall into the unknown category.

\textbf{Mechanism Allocation.}
A model may use different retrieval mechanisms across different samples. We therefore repeat the counterfactual patching procedure over many synthetic samples from a task and estimate the rate of each outcome as the fraction of samples classified as positional ($P$), lexical ($L$), reflexive ($R$), unknown, or no effect. We refer to these estimated rates as the model's mechanism allocation. We refer to the mechanism with the highest rate among $P$, $L$, and $R$ as the model's dominant mechanism.

Given this mechanism allocation, our primary comparison is between positional and semantic retrieval. We group the lexical and reflexive mechanisms as semantic mechanisms, since both retrieve based on entity information, and compare them to the positional mechanism which retrieves based on location. We define the positional ratio as \(\rpos = \frac{P}{L+R}\). Higher \rpos{} indicates a more positional mechanism allocation, while lower \rpos{} indicates a more semantic mechanism allocation. We use the mechanism allocation and positional ratio throughout this work to understand the internal mechanisms guiding in-context retrieval.
\section{Positional Encoding Choice Shapes Mechanism Allocation}
\label{sec:mechanism}
We use this mechanistic framework to understand how positional encoding choice shapes the retrieval mechanisms a model learns to use. We first show across 22 open-weight instruction-tuned models that PE hybrids exhibit a semantic shift compared to RoPE (\Cref{sec:broader_analysis}). We then study a controlled progression of pre-training ablations on PE choice (from RoPE to SWA RoPE to SWA NoPE), finding that SWA alone does not produce the shift, while additionally removing RoPE from global layers does (\Cref{sec:pretrain_ablation}). Finally, we find that restricting RoPE to local layers reduces the quality of positional information represented by the model, providing a potential explanation for the semantic shift (\Cref{sec:explanation}).

\begin{table}[t]
\centering
\caption{\textbf{Model families and descriptions.} We compare RoPE architectures that apply RoPE consistently throughout the model with PE hybrid architectures that vary the application of RoPE across layers. We report the PE configuration, as well as window size and local:global ratio where applicable. $^{\dagger}$ Gemma-4 2B/4B have a window size of 512 and Gemma-4 2B uses a 4:1 local:global ratio.}
\resizebox{\columnwidth}{!}{%
\begin{tabular}{lccc}
\toprule
\textbf{Model} & \textbf{PE Configuration} & \textbf{Window Size} & \textbf{Local:Global} \\
\midrule
\rowcolor{ropeblue!15} \multicolumn{4}{c}{\large\textbf{\textit{RoPE Architectures}}} \\
\rowcolor{ropeblue!15} Gemma-2 \citep{team2024gemma}    & RoPE (10K)  & 4096 & 1:1 \\
\rowcolor{ropeblue!15} Llama-3.1 \citep{grattafiori2024llama}  & RoPE (500K) & ---  & --- \\
\rowcolor{ropeblue!15} Qwen-2.5 \citep{qwen25}   & RoPE (1M)   & ---  & --- \\
\rowcolor{ropeblue!15} Qwen-3 \citep{yang2025qwen3}   & RoPE (1M)   & ---  & --- \\
\midrule
\rowcolor{hybridsalmon!15} \multicolumn{4}{c}{\large\textbf{\textit{PE Hybrid Architectures}}} \\
\rowcolor{hybridsalmon!15} Gemma-3 \citep{kamath2025gemma}       & RoPE (10K) / RoPE (1M)   & 1024 & 5:1 \\
\rowcolor{hybridsalmon!15} Gemma-4 \citep{gemma4team2026}$^{\dagger}$ & RoPE (10K) / p-RoPE (1M) & 512\textbar1024 & 4:1\textbar5:1 \\
\rowcolor{hybridsalmon!15} SmolLM3 \citep{bakouch2025smollm3}    & RoPE (5M) / NoPE         & ---  & 3:1 \\
\rowcolor{hybridsalmon!15} Command R7B \citep{yang2025hybrid}      & RoPE (50K) / NoPE        & 4096 & 3:1 \\
\bottomrule
\end{tabular}%
}
\label{tab:models}
\end{table}

\subsection{Model Battery}
\label{sec:broader_analysis}
In this section, we analyze mechanism allocation across eight model families to examine how PE choice shapes the learned mechanisms for in-context retrieval. We divide these models into the following two groups based on their PE configuration (\Cref{tab:models}):
{\setlength{\fboxsep}{1pt}
\begin{itemize}[leftmargin=*, labelsep=4pt, itemindent=0pt, itemsep=2pt, parsep=0pt, topsep=0pt]
    \item \fcolorbox{ropeblue!50}{ropeblue!15}{\textbf{\textit{RoPE Architectures.}}} Models that use RoPE \citep{rope} at every attention layer to encode positional information. Prior analyses of in-context retrieval mechanisms \citep{prakash2025language,gur2025mixing} studied only models of this type.
    \item \fcolorbox{hybridsalmon!50}{hybridsalmon!10}{\textbf{\textit{PE Hybrid Architectures.}}} We adopt a broad definition of PE hybrid in this work, considering any model that varies how positional encodings are applied across layers. These designs often combine this variation in PE with local and global attention layers. For example, Gemma-3 varies the RoPE base across local and global layers, Gemma-4 varies the RoPE base and uses p-RoPE \citep{barbero2024round} in global layers, and Command R7B and SmolLM3 both use NoPE layers. \Cref{tab:models} summarizes both the PE configuration and local/global attention structure of each architecture. Window size indicates the sliding window size \citep{beltagy2020longformer} where applicable.
\end{itemize}
}

These PE hybrids reflect a broader architectural trend away from applying RoPE uniformly throughout the model and toward differentiating PE across local and global layers. Very recent architectures, including Llama 4 \citep{meta25llama4} and MAI-Thinking-1 \citep{mai_thinking_1}, continue this trend. Our model battery therefore lets us ask what impact this shift in PE design has on retrieval mechanisms.

\textbf{Experimental Details.} We evaluate 22 instruction-tuned models across eight families: thirteen RoPE models (Gemma-2 \{2B, 9B, 27B\}, Llama-3.1 \{8B, 70B\}, Qwen-2.5 \{3B, 7B, 32B, 72B\}, Qwen-3 \{1.7B, 4B, 8B, 14B\}), and nine PE hybrid models (Gemma-3 \{4B, 12B, 27B\}, Gemma-4 \{2B, 4B, 12B, 31B\}, Command R7B, SmolLM3-3B). We evaluate each model on all 29 entity binding tasks from \citet{gur2025mixing}. For each model, we perform the mechanism analysis from \Cref{sec:mechanistic_lens} with 1000 samples per task at its selected retrieval layer (\Cref{tab:layer_selection}, Appendix~\ref{app:layer_scan_sweep}). We classify each prediction as positional, lexical, reflexive, or unknown, pool predictions across tasks, and compute $\rpos = \frac{P}{L+R}$ from the pooled counts. See Appendix~\ref{app:detail_battery} for additional details.

\textbf{PE Hybrids Shift Mechanisms.} We report \rpos{} and mechanism fractions normalized over \(P\), \(L\), and \(R\) in \Cref{fig:mechanism_shift}, with the full outcome rates in Appendix~\ref{app:results_sweep}. We find that PE hybrids exhibit a systematic shift toward semantic retrieval. While every RoPE model has $\rpos \geq 0.59$ (Llama3.1 8B), every PE hybrid has $\rpos \leq 0.53$ (Command R7B). At the extremes, the positional mechanism accounts for $65\%$ of predictions in Qwen2.5 72B ($\rpos=1.85$), compared to only $22\%$ in Gemma-4 2B ($\rpos=0.28$), with semantic mechanisms correspondingly accounting for $35\%$ and $78\%$. Despite substantial variation in model family, parameter count, context length, and training data, we find a clear semantic shift going from RoPE to PE hybrid architectures.

Prior mechanistic analyses of in-context retrieval studied RoPE models \citep{prakash2025language,gur2025mixing}, and our extension to PE hybrids reveals a different regime. The positional mechanism accounts for the largest share among \(P\), \(L\), and \(R\) for every RoPE model we test, while for every PE hybrid, either the lexical or reflexive mechanism has the largest share (\Cref{tab:full_distribution}). In other words, the dominant mechanism consistently changes from positional to semantic across the two architecture classes. Although PE hybrids use the same retrieval mechanisms identified by prior work, they exhibit a qualitatively different mechanism allocation.

Our model battery reveals a clear difference in mechanism allocation between RoPE and PE hybrid models despite substantial variation in model family, scale, and architecture. However, this variation also makes it difficult to determine which architectural differences drive the shift toward semantic retrieval. We turn to a controlled set of pre-training ablations to isolate the effect of PE choice.

\begin{figure}[t]
\centering
\includegraphics[width=\textwidth]{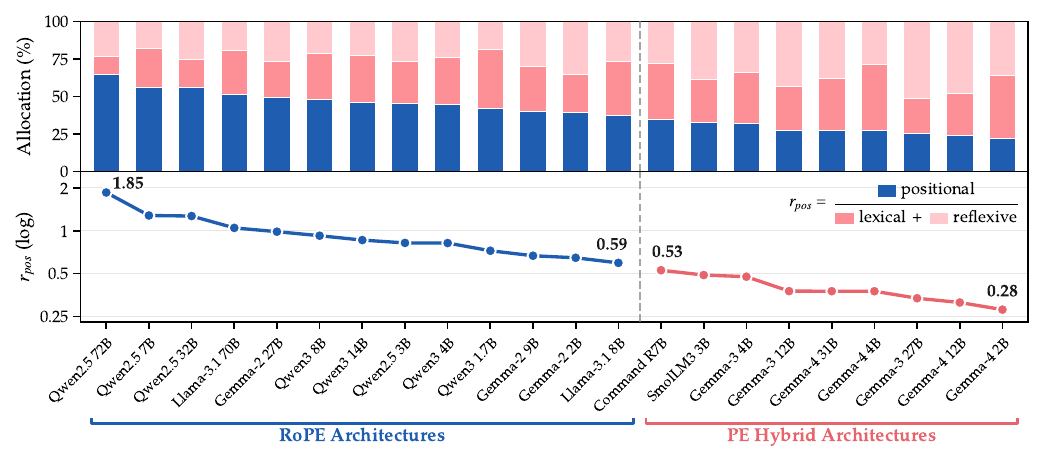}
\caption{Mechanism allocation across the model battery pooled over 29 tasks and sorted by \rpos{}. (\textbf{Top}) Relative allocation to positional, lexical, and reflexive retrieval mechanisms. (\textbf{Bottom}) Positional ratio for each model. The architecture classes separate completely: every RoPE model has a higher \rpos{} ($\geq 0.59$) than every PE hybrid ($\leq 0.53$).}
\label{fig:mechanism_shift}
\end{figure}

\begin{table}[t]
\centering
\caption{Mechanism allocation of three model architectures: RoPE, SWA RoPE, and SWA NoPE at two training checkpoints (16K and 32K training length) on the Boxes task. We report the percentage (\%) of samples classified as positional, lexical, reflexive, and unknown. We additionally report the ratio of positional to semantic (lexical, reflexive) mechanisms. At both training lengths, introducing SWA RoPE alone does not produce the semantic shift, but SWA with NoPE global layers substantially shifts the mechanism allocation, reducing \rpos{} by a factor of two.}
\label{tab:edenqiao-mechanism}
\begin{tabular}{llcccc!{\quad\vrule\quad}c}
\toprule
& Architecture & Positional & Lexical & Reflexive & Unknown & \rpos \\
\midrule
\multirow{3}{*}{16K}
 & RoPE     & 17.6 & 56.6 & 1.6 & 24.1 & 0.30 \\
 & SWA RoPE & 19.4 & 46.6 & 2.1 & 31.9 & 0.40 \\
 & SWA NoPE & 10.0 & 68.6 & 1.4 & 20.0 & \textbf{0.14} \\
\midrule
\multirow{3}{*}{32K}
 & RoPE     & 15.9 & 60.6 & 1.4 & 22.1 & 0.26 \\
 & SWA RoPE & 16.2 & 57.5 & 1.8 & 24.6 & 0.27 \\
 & SWA NoPE &  9.5 & 71.9 & 1.3 & 17.3 & \textbf{0.13} \\
\bottomrule
\end{tabular}
\end{table}

\subsection{Ablating Positional Encoding Choice}
\label{sec:pretrain_ablation}
To isolate whether positional encoding choice drives the shift in mechanism allocation, we utilize the controlled pre-training checkpoints from \citet{qiao2026rethinking}:
\begin{itemize}[leftmargin=*, labelsep=4pt, itemindent=0pt, itemsep=2pt, parsep=0pt, topsep=0pt]
    \item \textbf{RoPE.} Global attention for all layers, with RoPE \citep{rope} applied in every layer.
    \item \textbf{SWA RoPE.} Interleaving global attention and local sliding-window attention (SWA) \citep{beltagy2020longformer}, where tokens only attend to a fixed window size. RoPE still applied in every layer.
    \item \textbf{SWA NoPE.} The same alternating attention pattern as SWA RoPE, but RoPE is applied only in local layers, while global layers use NoPE \citep{puvvada2025swan, yang2025hybrid}.
\end{itemize}

Each architecture is first trained at 16K context length for 100B tokens, and then extended to 32K context length for an additional 5B tokens, totaling six checkpoints. These models each have 665M parameters and share the same training hyperparameters \citep[see][]{qiao2026rethinking}. Both SWA variants use a 1:1 ratio of local to global layers with a sliding window size of 128 tokens. We utilize the framework from \Cref{sec:mechanistic_lens} to classify each sample as positional ($P$), lexical ($L$), reflexive ($R$), or unknown, based on which token the model predicts under activation patching. We measure at layer 15 using 10K samples per model. Finally, since these are small base models, we evaluate on the Boxes task from \Cref{fig:background} to ensure reliable task performance. For further details see Appendix~\ref{app:detail_pretrain}.

We report our results in Table~\ref{tab:edenqiao-mechanism}. At both training lengths, SWA NoPE learns a substantially more semantic mechanism allocation than RoPE: \rpos{} falls from \(0.30\) to \(0.14\) for the 16K checkpoints and \(0.26\) to \(0.13\) for the 32K checkpoints. At both checkpoints, SWA NoPE reduces the positional mechanism and unknown rate, and increases the lexical mechanism.

In contrast, SWA RoPE does not exhibit this semantic shift and even increases \rpos{} at the 16K training checkpoint from \(0.30\) to \(0.40\), and nearly matches RoPE at 32K (\(0.26\) to \(0.27\)). The positional ratios are relatively low across all three architectures, consistent with the strong lexical bias of the Boxes task (see \Cref{fig:task_heatmap} and \Cref{app:t_entity} for more details). These controlled experiments show that SWA alone does not produce the semantic shift, while additionally removing RoPE from global layers does.

\subsection{Confining positional encoding degrades Ordering ID creation}
\label{sec:explanation}
Having established that the SWA NoPE architecture drives the shift toward semantic retrieval, we next investigate why this shift occurs. We first use our understanding of the three mechanisms guiding in-context retrieval to identify where PE choice could impact mechanism allocation. This analysis suggests that PE choice may affect how models form ordering IDs. We therefore examine whether SWA NoPE weakens the positional representations used to form ordering IDs.

\textbf{Where can PE shape mechanisms?}
To begin, we look at the three in-context retrieval mechanisms. The lexical and reflexive mechanisms are both semantic by design, storing either a semantic key into the target entity or a self-referential pointer to the target entity. Neither mechanism explicitly relies on positional information for retrieval, making the positional mechanism the natural candidate through which PE choice could directly affect mechanism allocation. We therefore dig deeper into how the positional mechanism operates.

Recent work studying in-context retrieval found that the positional mechanism occurs in two distinct stages via a lookback mechanism \citep{prakash2025language}. During the first stage, the model associates each answer with an ordering ID (OI) that represents its ordinal position in the list. For example, the answer token appearing third in the list stores a representation of its ordinal position (``third''). 

During retrieval, the model identifies the position specified by the query and matches it to the corresponding OI to recover the answer. Once created, the OI acts as a positional key for retrieval, analogous to the semantic key used by the lexical mechanism. This decomposition suggests that PE choice may affect positional retrieval by altering how these positional keys are formed. We therefore investigate whether OI creation differs across the same controlled pre-training ablations used above.

\textbf{SWA NoPE Confines PE.} Our controlled ablation showed that SWA NoPE shifts toward semantic retrieval, while SWA RoPE does not. Since ordering IDs encode position, their formation may depend on how positional information is provided by these architectures. SWA RoPE applies RoPE in both local and global layers, while SWA NoPE confines RoPE to local layers and uses NoPE in global layers. Next, we test whether confining PE to local layers makes OI creation more difficult.

\begin{wraptable}[16]{r}{0.55\columnwidth}
\vspace{-\baselineskip}
\centering
\small
\setlength{\tabcolsep}{4pt}
\caption{Linear probe accuracy (\%) on the residual stream at layer 15 across items in the list, predicting either the index in the list or the attached semantic key, across the six model checkpoints. Standard deviation reported in parentheses over eight runs. SWA NoPE has the worst ordering ID probe accuracy, while semantic key probe accuracy remains high.}
\label{tab:probe-oi}
\begin{tabular}{ll r@{\,}l r@{\,}l}
\toprule
Context & Model & \multicolumn{2}{c}{Ordering ID} & \multicolumn{2}{c}{Semantic Key} \\
\midrule
\multirow{3}{*}{16K}
 & RoPE     & 94.8\sd{0.2} &              & 99.0\sd{0.2} &            \\
 & SWA RoPE & 91.6\sd{0.1} & \dlt{-3.2}   & 98.8\sd{0.3} & \dlt{-0.2} \\
 & SWA NoPE & 82.7\sd{0.3} & \dlt{-12.1}  & 99.4\sd{0.1} & \dgt{+0.4} \\
\midrule
\multirow{3}{*}{32K}
 & RoPE     & 94.9\sd{0.2} &              & 99.1\sd{0.2} &            \\
 & SWA RoPE & 92.7\sd{0.2} & \dlt{-2.2}   & 99.0\sd{0.3} & \dlt{-0.1} \\
 & SWA NoPE & 83.9\sd{0.3} & \dlt{-11.0}  & 99.5\sd{0.1} & \dgt{+0.4} \\
\bottomrule
\end{tabular}
\end{wraptable}

\textbf{Experimental Details.}
To test whether ordering ID creation is impacted by PE choice, we train linear probes on the residual stream of the model. We use the same Boxes task from Section~\ref{sec:mechanism}, extracting the residual stream at each of the 20 answer tokens, and train a linear probe to predict the ordering ID (the entity's position in the list $1-20$). As a control, we train a linear probe to predict the semantic key (the object paired with the box), which checks that semantic information is still represented. We train probes at layer 15 since we found in \Cref{sec:pretrain_ablation} that this is where retrieval occurs, and extract the 1280-dimensional activations at this layer.

We fit an $L_2$-regularized logistic regression for each linear probe. We train on $1000$ prompts and test on a disjoint set of $1000$ prompts, forming training and test sets of $20\times1000=20{,}000$ samples each. We report the mean and standard deviation across eight independently sampled runs. We analyze the six checkpoints: RoPE, SWA RoPE, SWA NoPE, each trained at 16K context length and then extended to 32K. Additional details about the experimental setup can be found in Appendix~\ref{app:detail_probe}.

\textbf{SWA NoPE degrades OI creation.} Our results are shown in Table~\ref{tab:probe-oi}. Ordering IDs are substantially less linearly decodable under SWA NoPE, with probe accuracy decreasing by $12.1$ points at 16K and $11.0$ points at 32K compared to RoPE. SWA RoPE exhibits a much smaller decrease of $3.2$ and $2.2$ points, respectively. In contrast, semantic key accuracy remains nearly unchanged across all six checkpoints. As a result, the gap between semantic key and ordering ID accuracy grows from roughly $4$ points under RoPE to $16$ points under SWA NoPE. Overall, SWA NoPE selectively degrades the representation of OIs relative to semantic keys, consistent with its semantic shift.

\textbf{Mechanistic Explanation.} We hypothesize that confining PE to local layers impairs OI formation during training, shifting the model toward semantic retrieval. Under SWA NoPE, layers with explicit positional information have only a local view of the sequence, while layers with global context must infer position without explicit positional encoding, making globally consistent OIs more difficult to learn. Weaker OI formation may make positional retrieval less reliable, leading the model to rely more on semantic mechanisms whose underlying semantic representations remain intact.

This explanation predicts that as the sliding window shrinks relative to context length, OI formation should become more difficult and the mechanism shift should strengthen. Conversely, as the window approaches the full context length, the receptive field of the RoPE layers approaches the full context and this constraint should weaken. We leave testing this prediction with controlled checkpoints ablating window size for future work.
\section{Mechanism Allocation Shapes Long-Context Retrieval}
Having established in \Cref{sec:mechanism} that PE choice shifts the mechanisms models learn to use for retrieval, we next ask whether these differences predict long-context retrieval behavior. Prior work reports that SWA NoPE improves performance over RoPE \citep{yang2025hybrid, puvvada2025swan, qiao2026rethinking} on long-context benchmarks such as RULER \citep{hsieh2024ruler}. Given that SWA NoPE also shifts retrieval toward semantic mechanisms, these gains raise a natural question: do semantic retrieval mechanisms simply scale better to long contexts? We investigate this question by examining how performance varies across the RULER subtasks using the same pre-training checkpoints studied in \Cref{sec:mechanism}.

We re-evaluate the released checkpoints from \citet{qiao2026rethinking} on RULER \citep{hsieh2024ruler}. We evaluate both the 16K and 32K checkpoints at their respective training context lengths, using 500 samples per task. We evaluate on all thirteen tasks: eight needle-in-a-haystack (NIAH) tasks (three single key, three multi key, one multi value, and one multi query task), two QA tasks, a variable tracking task, and two aggregation tasks (common word extraction, frequent word extraction). We group the subtasks by retrieval setting to examine which tasks drive the aggregate RULER improvement. Additional experimental details can be found in Appendix~\ref{app:detail_behavior}.

\begin{table}[t]
\centering
\small
\caption{RULER scores across the three model architectures: RoPE, SWA RoPE, and SWA NoPE. The 16K checkpoints are trained for 100B tokens, and the 32K checkpoints extend them for a further 5B tokens, and each is evaluated at their training context length. SWA NoPE substantially improves the overall RULER score, with gains concentrated in most NIAH and QA tasks. However, it does not improve over RoPE on the first NIAH multi-key task or consistently improve on variable tracking and aggregation.}
\label{tab:ruler}
\begin{tabular}{llc|cc|cc}
\toprule
Context & Model & Overall (13) & NIAH mk1 (1) & VT + Aggr. (3) & NIAH rest (7) & QA (2) \\
\midrule
 \multirow{3}{*}{16K} & RoPE      & 47.1 & \textbf{91.2} & 16.9 & 56.1 & 38.7 \\
  & SWA RoPE  & 47.2 & 84.8 & \textbf{18.5} & 57.2 & 36.7 \\
  & SWA NoPE  & \textbf{58.3} & 78.0 & 17.5 & \textbf{77.2} & \textbf{43.4} \\
\midrule
 \multirow{3}{*}{32K} & RoPE      & 43.3 & \textbf{92.4} & 18.1 & 50.8 & 30.7 \\
  & SWA RoPE  & 45.7 & 81.6 & \textbf{23.0} & 55.1 & 28.8 \\
  & SWA NoPE  & \textbf{53.9} & 82.2 & 15.6 & \textbf{69.6} & \textbf{42.0} \\
\bottomrule
\end{tabular}
\end{table}

\begin{table}[t]
\centering
\small
\caption{We test each checkpoint on our confusable key task and compare to the RULER multi-value and multi-query subtasks. SWA NoPE struggles to select amongst competing keys, especially as they become more semantically similar, flipping the model ranking compared to multi-item NIAH where all keys are returned.}
\label{tab:confusable}
\begin{tabular}{llcccc|cc}
\toprule
& & \multicolumn{4}{c|}{Confusable keys} & \multicolumn{2}{c}{Multi-item} \\
\cmidrule(lr){3-6} \cmidrule(lr){7-8}
Context & Model & $0/4$ & $1/4$ & $2/4$ & $3/4$ & \textsc{multivalue} & \textsc{multiquery} \\
\midrule
 \multirow{3}{*}{16K} & RoPE      & \textbf{97.0} & \textbf{90.8} & \textbf{84.6} & \textbf{70.8} & 35.0 & 35.0 \\
  & SWA RoPE  & 92.6 & 80.4 & 75.4 & 60.2 & 44.0 & 38.4 \\
  & SWA NoPE  & 88.2 & 73.0 & 57.4 & 52.6 & \textbf{80.8} & \textbf{85.7} \\
\midrule
 \multirow{3}{*}{32K} & RoPE      & \textbf{96.8} & \textbf{91.8} & \textbf{82.6} & \textbf{67.6} & 25.2 & 30.5 \\
  & SWA RoPE  & 82.6 & 75.6 & 57.4 & 46.8 & 26.4 & 37.9 \\
  & SWA NoPE  & 76.0 & 67.6 & 46.8 & 38.6 & \textbf{53.8} & \textbf{65.7} \\
\bottomrule
\end{tabular}
\end{table}

\textbf{SWA NoPE is not uniformly better.} We report the RULER results in Table~\ref{tab:ruler}. Consistent with prior work \citep{qiao2026rethinking}, SWA NoPE substantially improves the overall RULER score over RoPE and SWA RoPE. However, this aggregate gain is not uniform across subtasks. SWA NoPE substantially improves most needle-in-a-haystack (NIAH) and QA tasks, but does not improve over RoPE on the first NIAH multi-key task or the variable tracking and aggregation tasks. We report the full task breakdown in Appendix~\ref{app:ruler}. These differences suggest that SWA NoPE changes which retrieval settings the model handles well rather than uniformly improving long-context retrieval.

\textbf{Mechanism Shift Predicts a Retrieval Trade-off.} The semantic shift identified in \Cref{sec:mechanism} suggests that SWA NoPE may excel when retrieving relevant content without discrimination among competing keys, but struggle when such discrimination is required. This distinction is reflected in the RULER results: SWA NoPE performs particularly well on multi-value and multi-query NIAH, where the requested values should all be returned rather than selecting one target while rejecting competing keys. To test the other side of this trade-off directly, we construct a retrieval task in which each context contains four needles, one queried key and three distractor keys, with each key consisting of four hyphenated words (e.g., \textit{daughter-others-prop-sudo}). Importantly, even when the keys share no words (\(0/4\)), retrieval requires distinguishing the queried key from the three distractors. We increase semantic confusability by varying the number of words shared across keys from zero to three.

We report the results in Table~\ref{tab:confusable}. Requiring the model to distinguish among competing keys reverses the model ordering even when the keys share no words (\(0/4\)): RoPE outperforms SWA NoPE, whereas SWA NoPE substantially outperforms RoPE on multi-value and multi-query retrieval. Increasing the semantic similarity among these competing keys further amplifies this difference, as predicted by the shift toward semantic retrieval. At 16K, the gap between RoPE and SWA NoPE grows from 8.8 points at \(0/4\) to 27.2 points at \(2/4\), with the same pattern at 32K, before narrowing at \(3/4\) as performance degrades across all architectures. Together, these results show that SWA NoPE's long-context gains reflect a shift in retrieval strategy with distinct strengths and failure modes rather than a uniform improvement in retrieval ability.
\section{Conclusion}
\label{sec:Conclusions}
In this work we analyzed how positional encoding (PE) choice shapes the in-context retrieval mechanisms a language model learns during training. Across controlled pre-training checkpoints and a battery of open-weight instruction-tuned models, we find that PE choice substantially shifts whether models rely on positional or semantic retrieval. Our controlled analysis further links this shift to degraded ordering ID representations when positional encoding is confined to local layers. These differences in mechanism allocation translate into different long-context retrieval profiles, rather than a uniform improvement or degradation in retrieval ability. As recent open-weight models increasingly adopt architectures that vary PE across layers, understanding these mechanistic consequences becomes increasingly important.

\textbf{Intentional Design.}
Through our mechanistic analysis we showed where and why SWA NoPE improves or degrades long-context retrieval, highlighting a broader role for interpretability in architectural design \citep{orgad2026interpretability}. By allowing model developers to better understand how architectural choices shape the internal mechanisms learned by the model, mechanistic analysis can help developers shape for their desired downstream behavior. More broadly, our results suggest that mechanistic interpretability can serve not only to explain existing models, but also as a tool for intentional design.

\textbf{Avoiding Long-Range RoPE.}
Recent work has highlighted theoretical limitations of RoPE at long context, leading to either token or positional confusion \citep{liu2026rotary, du2026rope}. Long-context extensions based on RoPE scaling, such as YaRN \citep{peng2024yarn}, additionally modify the positional encoding to extrapolate beyond the context lengths seen during training. Instead, the SWA NoPE architecture restricts RoPE to fixed-size sliding-window layers, and so the maximum positional distance remains fixed as the global context length increases. At the same time, our results show that confining positional encoding in this way substantially changes the model's learned retrieval mechanisms. We leave exploring the benefits and drawbacks of these choices to future work.

\section*{Limitations and Future Work}
\label{sec:Limitations}

\textbf{Scaling.} One limitation of this work is the cost associated with training models from scratch. While we are able to test small-scale checkpoints ablating the positional encoding choice \citep{qiao2026rethinking}, we are unable to train larger models, exhaustively ablate combinations of design choices, or analyze how the mechanisms develop during training. We leave these directions for future work.

\textbf{Data.} Mechanism allocation depends on which retrieval mechanisms are reinforced by the training objective, and therefore on the training data distribution. It would be interesting to study how different data distributions shape the in-context retrieval mechanisms learned by the model, or whether particular mechanisms could be targeted directly through the choice of long-context training data.

\textbf{Linear hybrids.} Many recent open-weight models have also adopted linear hybrids instead of PE hybrids \citep{team2025kimi, blakeman2025nvidia, team2026qwen35}. We provide a preliminary analysis of these models in Appendix~\ref{app:linear}, but leave a more comprehensive study for future work. In particular, it would be interesting to understand whether alternative efficient attention strategies shift the allocation of the same retrieval mechanisms studied here, or lead models to learn fundamentally different retrieval mechanisms.

\newpage
\section*{AI Statement}
In this work, we used generative AI tools for implementing methods in code and feedback on research methodology. Generating data sets, developing theoretical models, formulating/proving mathematical claims are not applicable to this work. Additionally, we used generative AI tools for help with plot and figure creation, as well as polishing the writing of the paper. We have reviewed all AI-assisted work, and stand by the research contributions made in this work. We take full responsibility for the final content of this work, including text, claims or artifacts produced with the aid of generative AI.

\section*{Acknowledgements}
We thank Linxi Zhao and Sofian Zalouk for insightful discussions on the ideas and manuscript. This research was supported by a gift to the LinkedIn–Cornell Bowers Strategic Partnership, and ARO grant W911NF-25-1-0254. Any opinions, findings, and conclusions or recommendations expressed in this material are those of the authors and do not necessarily reflect those of the sponsors.

\newpage

\bibliography{iclr2027_conference}
\bibliographystyle{iclr2027_conference}

\appendix
\newpage
\section{Experimental Details}
\label{app:experimental_details}

In this section, we describe the experimental details for our main experiments. We first describe the mechanistic lens used throughout this work (\Cref{app:detail_lens}), then the model battery (\Cref{app:detail_battery}), the pre-training ablation (\Cref{app:detail_pretrain}), linear probing (\Cref{app:detail_probe}), and long-context behavior (\Cref{app:detail_behavior}). All experiments are run in bf16 on a single NVIDIA RTX 6000 Ada (48GB), except models larger than 25B parameters, which exceed its memory and are run on an NVIDIA B200.

\subsection{Mechanistic Lens}
\label{app:detail_lens}
In this section we describe general details about the framework we use for our main mechanistic analysis. We utilize the datasets and framework from \citet{gur2025mixing} that disentangles three mechanisms that language models use to perform in-context retrieval: positional, lexical, and reflexive. In total they propose 29 datasets: ten synthetic groupings (Filling Liquids, People and Objects, Programming Dictionary, Music, Biology Experiment, Chemistry Experiment, Transportation, Sports Events, Space Observations, Boxes). While the Boxes task only has two entities (Object, Box), each other task queries up to three different entities (first, second, third), totaling 29 task/query pairs.

The given LM is run on two prompts: the counterfactual prompt and original prompt. Then, at the given layer $\ell$, the activations from the LM run on the counterfactual prompt are patched to the LM run on the original prompt. Through this patching procedure, each of the three mechanisms predicts a different token in the original prompt (see \citet{gur2025mixing} for additional details on how the mechanisms are disentangled). Across 1000 generated pairs, the mechanism allocation can be measured by counting the number of samples that follow each of the mechanisms.

We prompt at multiple tokens for all our results, and patch at a different layer for each model. The full breakdown for our battery is shown in \Cref{tab:layer_selection} highlighting our choices for each model.

\begin{table}[ht]
\centering
\small
\begin{tabular}{lcc|lcc}
\toprule
\multicolumn{3}{c}{\textbf{RoPE}} & \multicolumn{3}{c}{\textbf{PE Hybrid}} \\
\textbf{Model} & \textbf{Layer} & \textbf{Positions} & \textbf{Model} & \textbf{Layer} & \textbf{Positions} \\
\midrule
Gemma-2 2B    & 18 & $-1,-4,-6,-8$   & Gemma-3 4B  & 23 & $-1,-4,-6,-8$ \\
Gemma-2 9B    & 28 & $-1,-4,-6,-8$   & Gemma-3 12B & 29 & $-1,-4,-6,-8$ \\
Gemma-2 27B   & 25 & $-1,-4,-6,-8$   & Gemma-3 27B & 41 & $-1,-4,-6,-8$ \\
Qwen2.5 3B    & 31 & $-1,-4,-6,-8$   & Gemma-4 2B & 24 & $-1,-4,-6,-8$ \\
Qwen2.5 7B    & 22 & $-1,-4,-6,-8$   & Gemma-4 4B & 29 & $-1,-4,-6,-8$ \\
Qwen2.5 32B   & 51 & $-1,-4,-6,-8$   & Gemma-4 12B & 35 & $-1,-8,-10,-12$ \\
Qwen2.5 72B   & 62 & $-1,-4,-6,-8$   & Gemma-4 31B & 41 & $-1,-8,-10,-12$ \\
Llama-3.1 8B  & 16 & $-1,-4,-6,-8$   & Command R7B & 19 & $-1,-3,-5,-7$ \\
Llama-3.1 70B & 35 & $-1,-4,-6,-8$   & SmolLM3 3B  & 23 & $-1,-4,-6,-8$ \\
Qwen3 1.7B    & 19 & $-1,-8,-10,-12$ &             &    & \\
Qwen3 4B      & 24 & $-1,-8,-10,-12$ &             &    & \\
Qwen3 8B      & 24 & $-1,-8,-10,-12$ &             &    & \\
Qwen3 14B     & 28 & $-1,-8,-10,-12$ &             &    & \\
\bottomrule
\end{tabular}
\caption{Patched layer and token positions per model. Positions are matched by token role: the `?' and `:' of the query, plus the final two chat-scaffold positions. Offsets differ with template length.}
\label{tab:layer_selection}
\end{table}

\subsection{Model Battery}
\label{app:detail_battery}
In this section we describe details about the larger model battery. We evaluate 22 different post-trained models. RoPE: Qwen2.5 (3B, 7B, 32B, 72B), Qwen3 (1.7B, 4B, 8B, 14B), Llama-3.1 (8B, 70B), and Gemma-2 (2B, 9B, 27B). PE hybrid: Gemma-3 (4B, 12B, 27B), Gemma-4 (2B, 4B, 12B, 31B), Command R7B, and SmolLM3 3B. We also average across the full task battery using 20 entities, and evaluate with 1000 samples for each task. Two important decisions for each model are the choice of positions to patch, as well as the layer selection. Since these are all instruction-tuned models with their own chat templates, we match the patched tokens by role (final token, newline, `:', `?') and perform layer selection in \Cref{app:layer_scan_sweep} to select these for each model. Our results are shown in \Cref{tab:layer_selection}. Gemma-2, Qwen2.5, and Llama3.1 follow prior work, while we compute the selections for the new models ourselves.

\subsection{Pre-training Ablation}
\label{app:detail_pretrain}
In this section we describe details about the model checkpoints from the pre-training ablation \citep{qiao2026rethinking}, and our mechanistic analysis on these models.

We evaluate these models on Boxes because this task is relatively simple and they are 665M parameter base models that struggle on most tasks. Additionally, we use a cloze format (appending `The object is in Box' to the end of the prompt) to ensure these models can retain high accuracy on this task. We run 10K samples per model to get a robust measure of the mechanism allocation, use only the last token position for patching, and patch at layer 15. The 16K checkpoints were trained for 100B tokens at 16K context length, and the 32K checkpoints were trained for an additional 5B tokens at 32K context length. \citet{qiao2026rethinking} trains three different architectures: RoPE, SWA RoPE, and SWA NoPE. Both SWA architectures use a sliding window size of 128 tokens, with a ratio of 1:1 local:global layers. Further, SWA NoPE removes RoPE from the global layers but retains RoPE in the sliding window layers. These models also use an attention sink.

\begin{figure}[h]
\centering
\includegraphics[width=\textwidth]{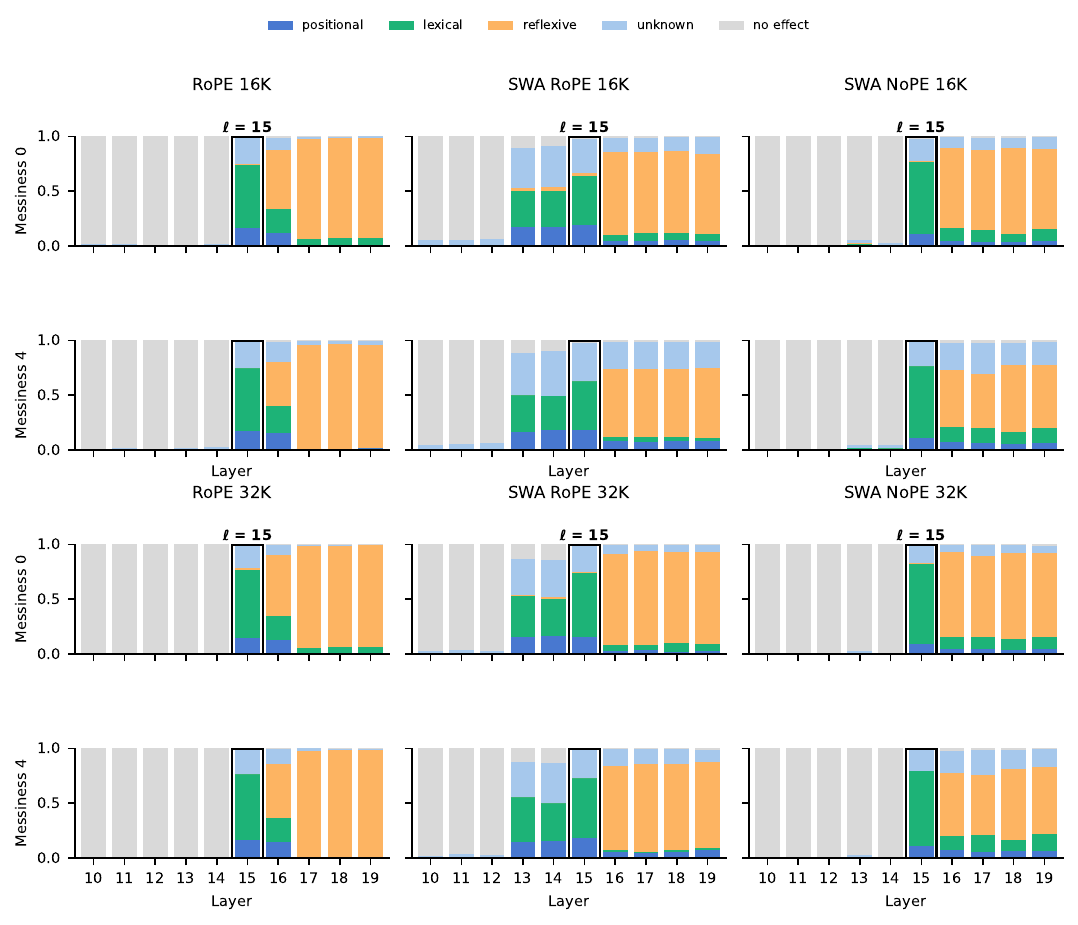}
\caption{For every checkpoint in our pre-training ablation, layer 15 is the last layer before the answer is retrieved.}
\label{fig:layer_scan_pretrain}
\end{figure}

To validate our choice of layer 15, we present layer scans for these models. Our results are shown in \Cref{fig:layer_scan_pretrain}. We show that layer 15 is the last layer before the answer is retrieved for each model in our pre-training checkpoints. The messiness 4 setting differentiates the reflexive self-referential pointer from the counterfactual prompt's answer by ensuring the object does not exist in the original prompt, so it can only be reflexive in messiness 4 if the answer has already been retrieved. See \citet{gur2025mixing} for additional details on the layer-selection procedure.

\subsection{Linear Probing}
\label{app:detail_probe}
In this section we provide additional details for the linear probing experiments.

We probe the three pre-training checkpoints (RoPE, SWA RoPE, SWA NoPE), each with 30 layers and $d_{\text{model}}=1280$. We evaluate both the 16K checkpoints trained for 100B tokens and the corresponding 32K checkpoints extended for an additional 5B tokens. For all models, we probe at layer 15, the same layer used for our mechanism analysis. We extract the residual stream entering block 15 using a forward pre-hook, with no KV cache.

We use the Boxes task with $N=20$ entities (``the \{Object\} is in Box \{Box\}''). For each entity, we extract the residual at the Box-letter token. For the \emph{ordering ID} probe, the label is the entity's position in the list ($20$ classes, chance $5\%$). For the \emph{Semantic Key} control, we predict the object bound to each box ($82$ classes, chance $\approx1.2\%$). Objects are randomized across positions between prompts, making the semantic-key label independent of position.

We fit multinomial logistic regression with L2 regularization ($C=1$) on standardized features using scikit-learn (\texttt{StandardScaler} followed by \texttt{LogisticRegression}). We use all 1280 dimensions without dimensionality reduction. Each repeat uses 1000 training prompts (20{,}000 examples) and a separate set of 1000 held-out test prompts. We run eight repeats, redrawing both sets and refitting the probe each time, and report mean held-out accuracy $\pm$ SD across repeats.

\subsection{Long-Context Behavior}
\label{app:detail_behavior}
In this section we provide additional details for our RULER evaluation and confusable-key task. We evaluate our pre-training checkpoints on RULER following \citet{hsieh2024ruler}, using 500 samples per task. We evaluate the 16K checkpoints at 16K context length and the 32K checkpoints at 32K context length.

Our confusable-key task builds on NIAH-mk1, where we replace each key with four hyphenated words and vary the number of words shared across competing keys. Each prompt contains four key-value pairs inserted into a haystack of Paul Graham essays.

For a $k/4$ example, we sample $k$ words shared across all four keys and independently sample the remaining $4-k$ words for each key, with no overlap between the remaining words. We then independently shuffle the four words within each key before joining them with hyphens. Thus, $0/4$ contains four disjoint keys, while at $3/4$ all four keys share three words and differ by only one. Values are uniformly sampled 7-digit numbers, and key-value pairs use the standard RULER needle template. We uniformly sample one of the four keys to query, giving a chance level of $25\%$.

We score a prediction as correct when the target value appears in the model generation. We generate 500 samples at each similarity level ($0/4$, $1/4$, $2/4$, $3/4$) for each checkpoint at both 16K and 32K context lengths. The $0/4$ condition uses the same haystack, needle and query structure as NIAH-mk1, but replaces its single-word keys with disjoint four-word keys. It therefore controls for the change in key format while varying only key similarity across the $0/4$--$3/4$ ladder.

\newpage
\section{Layer Selection Validation}
\label{app:layer_scan_sweep}

\begin{figure}[h]
\centering
\includegraphics[width=.6\textwidth]{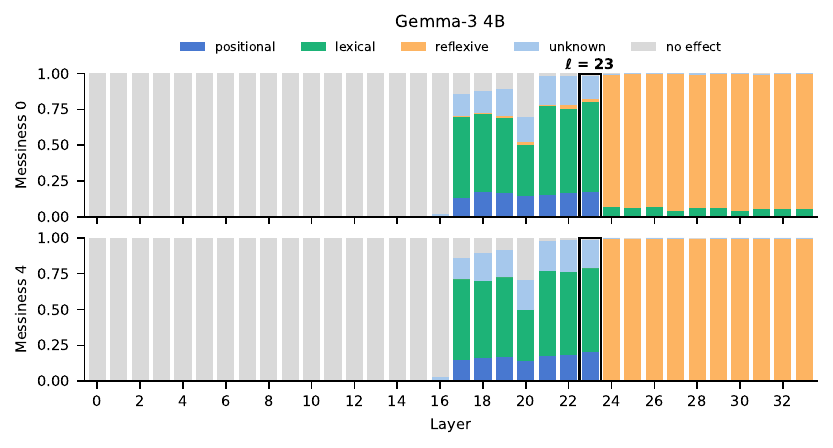}
\caption{Layer scan for Gemma-3 4B with the messiness-0 (top) and messiness-4 (bottom) counterfactuals. Bars show the share of each mechanism when patching at each layer.}
\label{fig:layer_scan_g34}
\end{figure}

\begin{figure}[h]
\centering
\includegraphics[width=.6\textwidth]{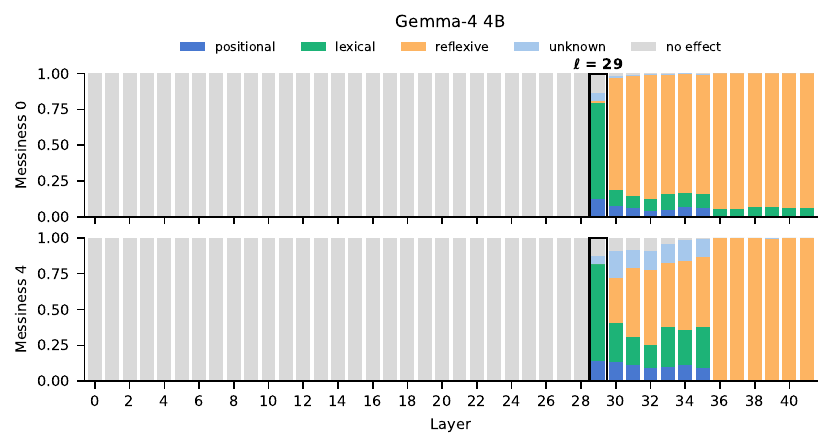}
\caption{Layer scan for Gemma-4 4B with the messiness-0 (top) and messiness-4 (bottom) counterfactuals. Bars show the share of each mechanism when patching at each layer.}
\label{fig:layer_scan_g44}
\end{figure}

\begin{figure}[ht]
\centering
\includegraphics[width=.6\textwidth]{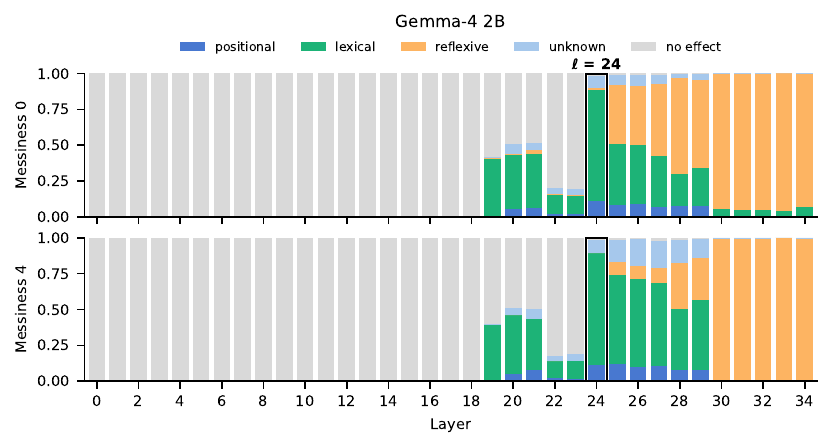}
\caption{Layer scan for Gemma-4 2B with the messiness-0 (top) and messiness-4 (bottom) counterfactuals. Bars show the share of each mechanism when patching at each layer.}
\label{fig:layer_scan_g42}
\end{figure}

In this section we describe the protocol we use to select the patching layer for our main results. We mostly follow the setup of \citet{gur2025mixing} and select the last layer at which patching does not transfer an already-retrieved answer. To test this, we use their messiness-4 counterfactual, in which the counterfactual answer is a new item that does not appear in the original prompt. If the model outputs this answer after patching, the answer can only have come from the patched activation, so retrieval has already happened by that layer. Scans patch the same positions as the main experiments and use the boxes task with query category 1. 
We show layer scans for three models to illustrate the process. 
\begin{itemize}[leftmargin=*, labelsep=4pt, itemindent=0pt, itemsep=2pt, parsep=0pt, topsep=0pt]
    \item Gemma-3 4B (\Cref{fig:layer_scan_g34}) has a clear shift from retrieval in layer 23 to almost entirely reflexive in layer 24 and the messiness-4 condition returns entirely the counterfactual answer, so we select $\ell=23$.
    \item Gemma-4 4B (\Cref{fig:layer_scan_g44}) is also clean: after layer 29 the messiness-4 scan shows a substantial share of samples returning the counterfactual answer, so we select $\ell=29$.
    \item Gemma-4 2B (\Cref{fig:layer_scan_g42}) is less clean: after layer 24 there is a small increase in the counterfactual answer. We therefore select $\ell=24$, erring on the side of preventing an already-retrieved counterfactual answer from entering the mechanism allocation analysis.
\end{itemize}

Our selected retrieval layer for all PE hybrids also ended up being the global layer when there was a mix of local and global layers. Gemma-4 2B, which has a 4:1 ratio of local to global layers (global layers 4, 9, 14, \ldots), illustrates this: patching first has an effect at global layer 19, the lexical share rises sharply at global layer 24, and the messiness-4 answer is fully retrieved from layer 30, immediately after global layer 29.

\section{Mechanism Allocation Extended Results}
\label{app:results_sweep}
In this section, we report the full mechanism distribution in \Cref{tab:full_distribution}: positional, lexical, reflexive, unknown, and no effect, along with the resulting $\rpos$. While the no-effect rate is generally low, it is substantially higher for some models, particularly Gemma-4 12B, even when patching multiple tokens.

\begin{figure}[ht]
\centering
\includegraphics[width=\textwidth]{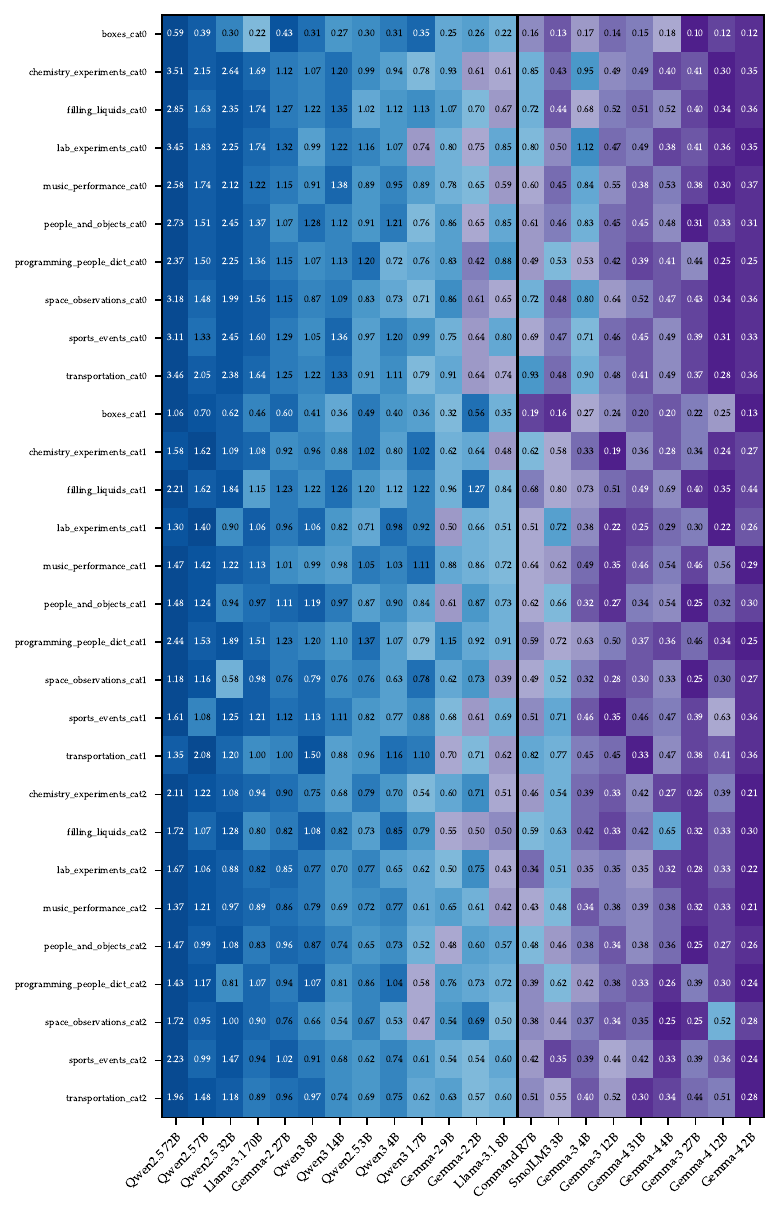}
\caption{Full per-task breakdown of $\rpos$ across the model battery (29 tasks). Separation does not hold for every individual task, but holds when averaged across tasks.}
\label{fig:task_heatmap}
\end{figure}

\Cref{fig:task_heatmap} shows $\rpos$ separately for each task. The ordering between RoPE and PE hybrids does not hold for every individual task, but emerges in aggregate. Across tasks, however, there is a strong shift in $\rpos$ from the most positional models to the most semantic models.

\subsection{Full Distribution Table}
\begin{table}[h]
\centering
\small
\begin{tabular}{lrrrrrr}
\toprule
\textbf{Model} & \textbf{Positional} & \textbf{Lexical} & \textbf{Reflexive} & \textbf{Unknown} & \textbf{No effect} & $r_{pos}$ \\
\midrule
\multicolumn{7}{l}{\textit{RoPE}} \\
Qwen2.5 72B & \textbf{46.0} & 8.4 & \underline{16.4} & 21.9 & 7.3 & 1.85 \\
Qwen2.5 7B & \textbf{36.0} & \underline{16.7} & 11.4 & 33.0 & 2.7 & 1.28 \\
Qwen2.5 32B & \textbf{37.9} & 12.8 & \underline{17.0} & 29.1 & 3.2 & 1.27 \\
Llama-3.1 70B & \textbf{28.8} & \underline{16.7} & 10.8 & 28.5 & 15.2 & 1.05 \\
Gemma-2 27B & \textbf{33.0} & 15.7 & \underline{17.8} & 25.1 & 8.2 & 0.98 \\
Qwen3 8B & \textbf{29.3} & \underline{18.9} & 12.9 & 35.4 & 3.5 & 0.92 \\
Qwen3 14B & \textbf{30.5} & \underline{20.5} & 15.0 & 30.1 & 3.9 & 0.86 \\
Qwen2.5 3B & \textbf{27.6} & \underline{17.2} & 16.4 & 36.4 & 2.3 & 0.82 \\
Qwen3 4B & \textbf{26.3} & \underline{18.1} & 14.1 & 38.0 & 3.5 & 0.82 \\
Qwen3 1.7B & \textbf{22.5} & \underline{21.2} & 10.0 & 43.2 & 3.2 & 0.72 \\
Gemma-2 9B & \textbf{29.1} & \underline{21.9} & 21.7 & 25.4 & 1.9 & 0.67 \\
Gemma-2 2B & \textbf{26.2} & 16.9 & \underline{23.6} & 31.1 & 2.2 & 0.65 \\
Llama-3.1 8B & \textbf{24.1} & \underline{23.1} & 17.4 & 27.4 & 8.0 & 0.59 \\
\midrule
\multicolumn{7}{l}{\textit{PE hybrid}} \\
Command R7B & \underline{21.8} & \textbf{23.8} & 17.5 & 30.8 & 6.1 & 0.53 \\
SmolLM3 3B & \underline{18.6} & 16.1 & \textbf{21.9} & 40.6 & 2.7 & 0.49 \\
Gemma-3 4B & 22.4 & \underline{23.5} & \textbf{23.7} & 28.5 & 1.9 & 0.48 \\
Gemma-3 12B & 21.9 & \underline{23.3} & \textbf{34.8} & 17.4 & 2.6 & 0.38 \\
Gemma-4 31B & 21.2 & \underline{26.9} & \textbf{29.3} & 15.0 & 7.6 & 0.38 \\
Gemma-4 4B & 17.6 & \textbf{28.4} & \underline{18.5} & 13.1 & 22.4 & 0.38 \\
Gemma-3 27B & \underline{20.9} & 19.7 & \textbf{42.5} & 14.5 & 2.4 & 0.34 \\
Gemma-4 12B & 12.7 & \underline{15.1} & \textbf{25.4} & 6.5 & 40.3 & 0.31 \\
Gemma-4 2B & 17.1 & \textbf{33.1} & \underline{28.0} & 18.8 & 3.0 & 0.28 \\
\bottomrule
\end{tabular}
\caption{Mechanism distribution per model, as a percentage of all samples (29 task--query pairs, 29{,}000 samples each), sorted by $r_{pos}=P/(L+R)$. \textbf{Bold}/\underline{underline}: highest/second-highest among Positional, Lexical, and Reflexive.}
\label{tab:full_distribution}
\end{table}

\clearpage
\subsection{Mechanism Allocation by Target Entity Position}
\label{app:t_entity}
\begin{figure}[ht]
\centering
\includegraphics[width=.8\textwidth]{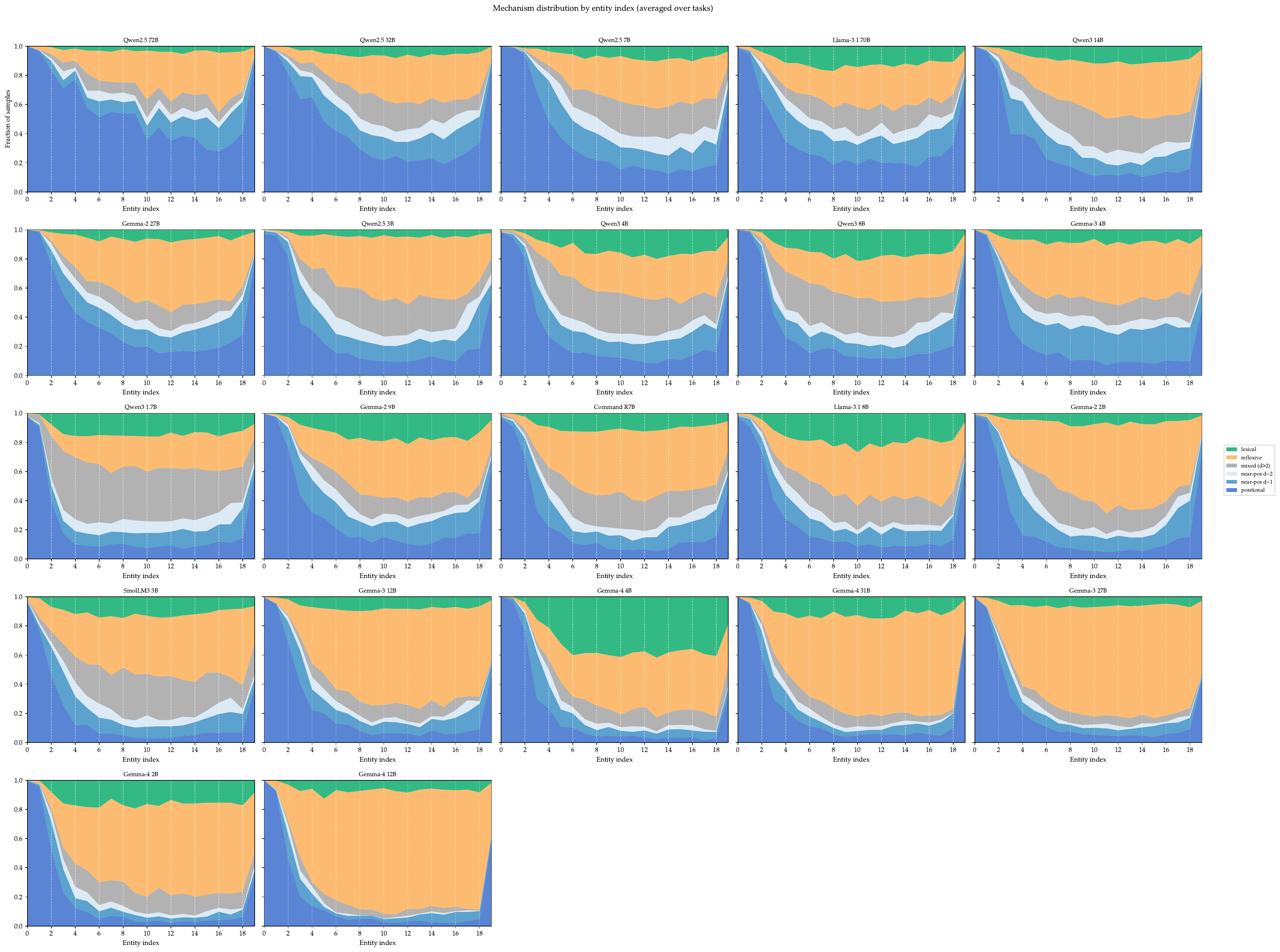}
\caption{Mechanism allocation by entity index for $t_{entity}=1$, averaged over tasks. Reflexive retrieval dominates lexical retrieval because the target entity always precedes the query entity.}
\label{fig:t_entity_cq0}
\end{figure}

\begin{figure}[ht]
\centering
\includegraphics[width=.8\textwidth]{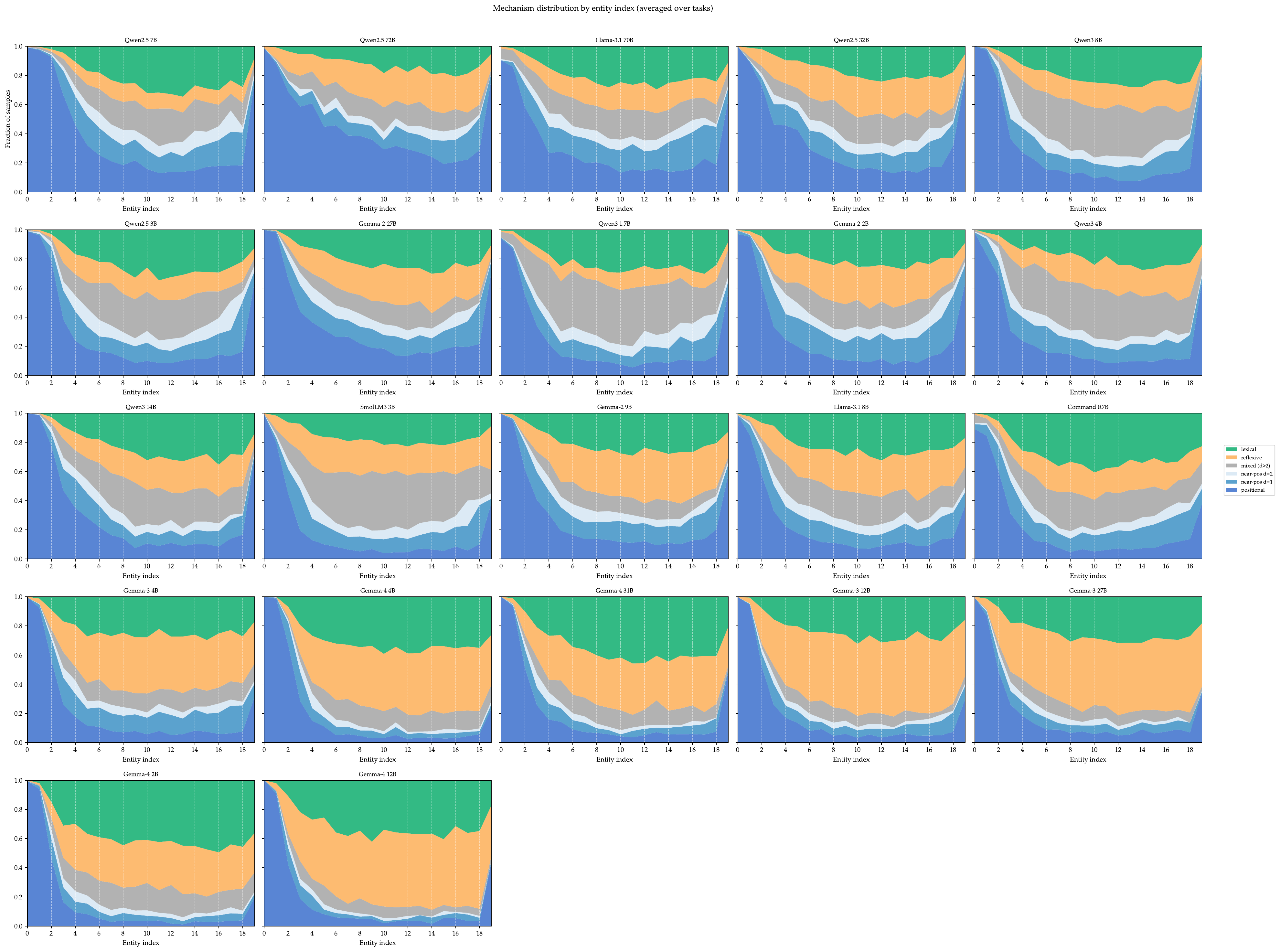}
\caption{Mechanism allocation by entity index for $t_{entity}=2$, averaged over tasks.}
\label{fig:t_entity_cq1}
\end{figure}

\begin{figure}[t]
\centering
\includegraphics[width=.8\textwidth]{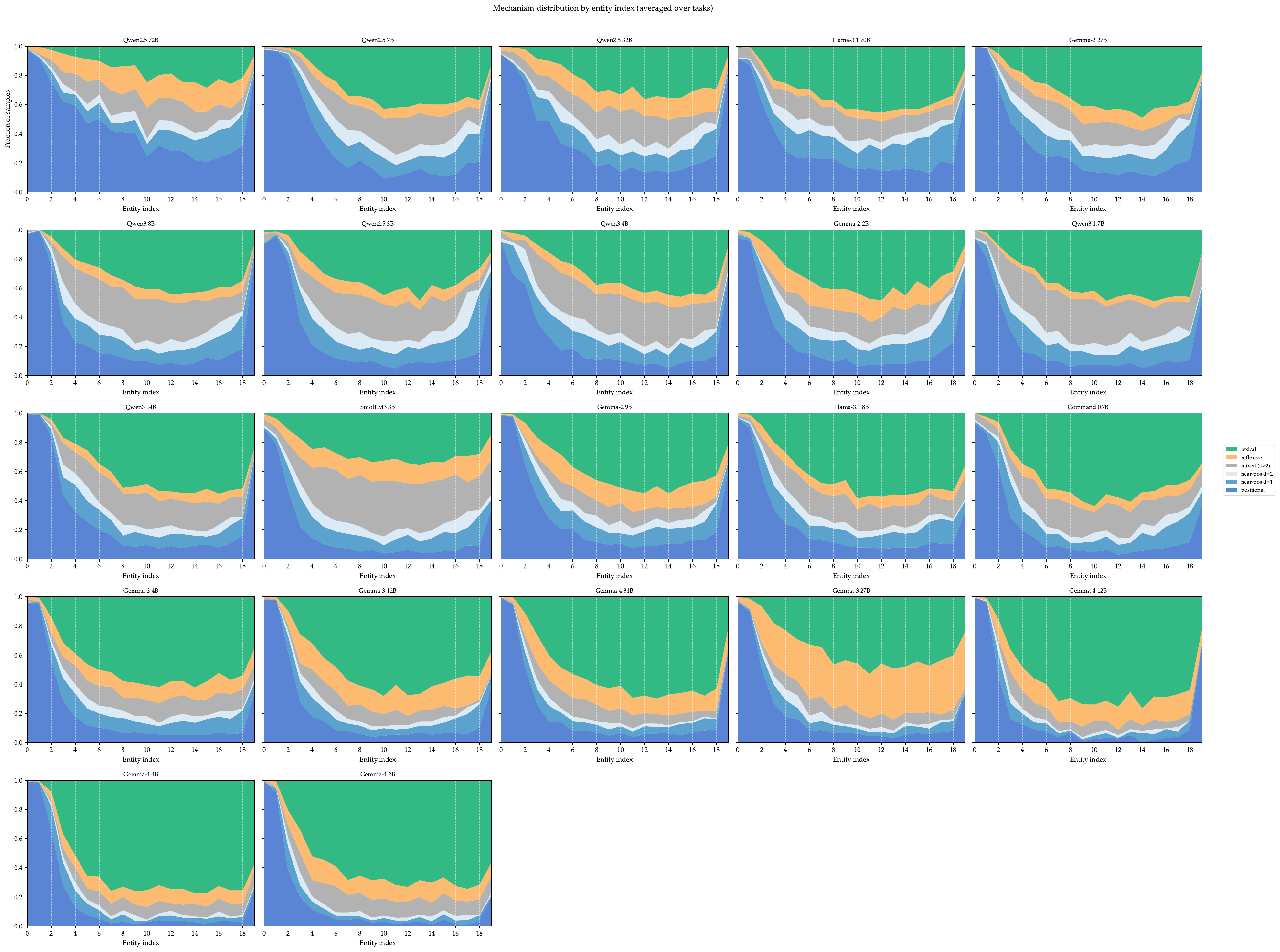}
\caption{Mechanism allocation by entity index for $t_{entity}=3$, averaged over tasks. Lexical retrieval dominates reflexive retrieval because the target entity follows the query entity.}
\label{fig:t_entity_cq2}
\end{figure}

\Cref{fig:t_entity_cq0,fig:t_entity_cq1,fig:t_entity_cq2} show mechanism allocation as a function of target entity position $t_{entity}$ for each model. As shown by \citet{gur2025mixing}, $t_{entity}$ controls the tradeoff between lexical and reflexive mechanisms: when the target precedes the query entity in the group, models shift toward reflexive retrieval. This breakdown confirms that our results hold consistently across all query types.

\section{RULER Extended Results}
\label{app:ruler}

In \Cref{tab:ruler_full}, we report the per-task RULER results for the pre-training ablation. The per-task results show that SWA NoPE's gains are concentrated rather than uniform. Its largest improvements occur on several NIAH tasks, particularly S3, MK3, MV, and MQ, as well as both QA tasks. However, SWA NoPE harms or does not improve performance on MK1 and the variable-tracking and aggregation tasks, motivating examining how the shift in retrieval mechanism interacts with the structure of the retrieval task.

\begin{table*}[h]
\centering
\small
\setlength{\tabcolsep}{3pt}
\caption{Per-task RULER scores at 16K and 32K context. $N=500$ per task. Best result for each task and context length is bolded.}
\label{tab:ruler_full}
\begin{tabular}{ll cccccccc ccccc}
\toprule
& & \multicolumn{8}{c}{NIAH} & \multicolumn{3}{c}{VT + Aggr.} & \multicolumn{2}{c}{QA} \\
\cmidrule(lr){3-10} \cmidrule(lr){11-13} \cmidrule(lr){14-15}
Context & Model & S1 & S2 & S3 & MK1 & MK2 & MK3 & MV & MQ & VT & CWE & FWE & QA1 & QA2 \\
\midrule
 \multirow{3}{*}{16K} & RoPE      & 88.2 & 97.4 & \textbf{50.0} & \textbf{91.2} & 83.0 & 4.0 & 35.0 & 35.0 & 0.1 & 0.6 & \textbf{50.1} & 50.6 & 26.8 \\
  & SWA RoPE  & \textbf{100.0} & 94.6 & 36.6 & 84.8 & 75.2 & 11.6 & 44.0 & 38.4 & \textbf{6.1} & \textbf{5.9} & 43.4 & 45.8 & 27.6 \\
  & SWA NoPE  & \textbf{100.0} & \textbf{100.0} & 48.0 & 78.0 & \textbf{84.2} & \textbf{41.6} & \textbf{80.8} & \textbf{85.7} & 0.2 & 3.5 & 48.7 & \textbf{54.8} & \textbf{32.0} \\
\midrule
 \multirow{3}{*}{32K} & RoPE      & \textbf{100.0} & 99.6 & 18.8 & \textbf{92.4} & \textbf{81.2} & 0.2 & 25.2 & 30.5 & 14.4 & 1.1 & 38.7 & 33.6 & 27.8 \\
  & SWA RoPE  & \textbf{100.0} & \textbf{99.8} & 27.2 & 81.6 & 74.0 & 20.2 & 26.4 & 37.9 & \textbf{14.9} & \textbf{9.1} & \textbf{45.0} & 30.8 & 26.8 \\
  & SWA NoPE  & \textbf{100.0} & 97.6 & \textbf{77.6} & 82.2 & 68.8 & \textbf{24.0} & \textbf{53.8} & \textbf{65.7} & 2.5 & 5.1 & 39.1 & \textbf{51.8} & \textbf{32.2} \\
\bottomrule
\end{tabular}
\end{table*}

In \Cref{tab:confusable} we showed that SWA RoPE degrades on the confusable-key task even though its mechanism allocation stays similar to RoPE. To study why this is, we categorize the incorrect predictions of each model at 32K context length. As shown in \Cref{tab:error_analysis}, 82-94\% of SWA NoPE's errors at 32K retrieve a competing key's value. In contrast, roughly half of SWA RoPE's errors return no needle value and instead consist primarily of degenerate repetitions. Thus, SWA RoPE's reduced accuracy at 32K reflects a distinct failure mode rather than the competing-key retrieval errors that characterize SWA NoPE.

\begin{table*}[ht]
\centering
\small
\setlength{\tabcolsep}{3pt}
\caption{Error types on the confusable-key task at 32K context length. Conditioned on an incorrect answer, we report the percentage of errors that return a value bound to one of the three competing keys (Competing key), versus no needle value (Neither).}
\label{tab:error_analysis}
\begin{tabular}{llcccccccc}
\toprule
& & \multicolumn{2}{c}{$0/4$} & \multicolumn{2}{c}{$1/4$} & \multicolumn{2}{c}{$2/4$} & \multicolumn{2}{c}{$3/4$} \\
\cmidrule(lr){3-4} \cmidrule(lr){5-6} \cmidrule(lr){7-8} \cmidrule(lr){9-10}
Context & Model & Comp. key & Neither & Comp. key & Neither & Comp. key & Neither & Comp. key & Neither \\
\midrule
 \multirow{3}{*}{32K} & RoPE     & 94\% & 6\% & 83\% & 17\% & 86\% & 14\% & 76\% & 24\% \\
  & SWA RoPE & 43\% & 57\% & 39\% & 61\% & 52\% & 48\% & 53\% & 47\% \\
  & SWA NoPE & 94\% & 6\% & 86\% & 14\% & 82\% & 18\% & 83\% & 17\% \\
\bottomrule
\end{tabular}
\end{table*}

\newpage
\section{Linear Hybrids}
\label{app:linear}
Many recent open-weight models increasingly use linear attention hybrids \citep{team2025kimi, blakeman2025nvidia, team2026qwen35}, often motivated by long-context gains that are primarily computational. A natural question is how these architectures affect in-context retrieval mechanisms.

We briefly examine three linear hybrids. Qwen3.5 4B interleaves Gated DeltaNet layers with full p-RoPE attention at a 3:1 ratio, RecurrentGemma 2B interleaves gated linear recurrence with local p-RoPE attention (window 2048) at a 2:1 ratio, and Granite-4.0-H 1B interleaves Mamba2 layers with full NoPE attention at a 9:1 ratio. We run the full task battery on each using the same protocol as the main experiments.

\begin{figure}[h]
\centering
\includegraphics[width=\textwidth]{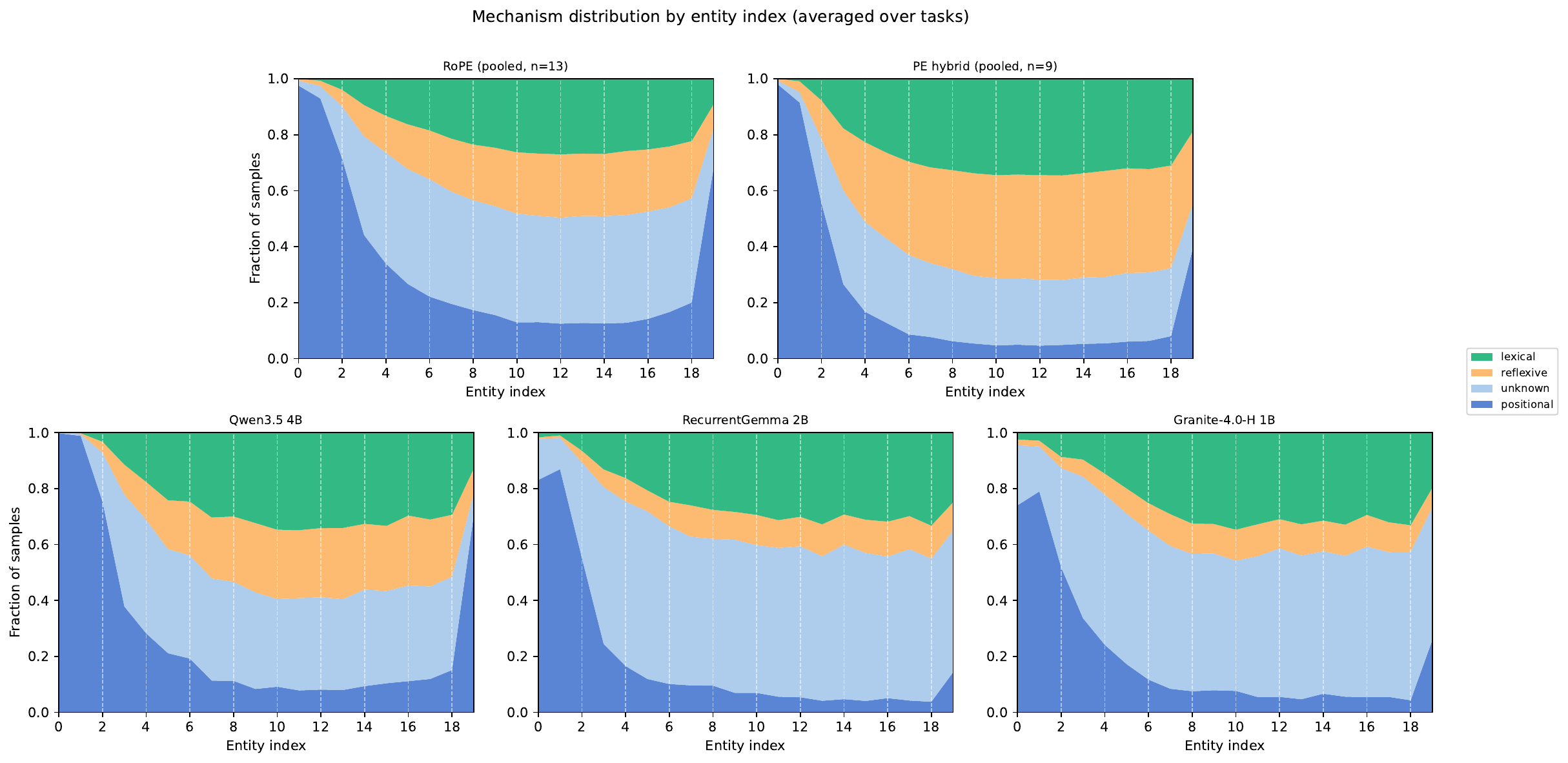}
\caption{Mechanism allocation across linear hybrids compared to RoPE/PE hybrids.}
\label{fig:linear_hybrids}
\end{figure}

\textbf{Results.} None of the three exhibits the mechanism allocation characteristic of the PE hybrids in our main battery. Qwen3.5 falls within the RoPE range ($\rpos=0.72$). RecurrentGemma ($0.59$) and Granite ($0.60$) instead lie near the boundary: roughly matching the lowest RoPE model, Llama-3.1 8B ($0.59$), while remaining above the highest PE hybrid, Command R7B ($0.53$). \Cref{fig:linear_hybrids} breaks down mechanism allocation by entity index. In RecurrentGemma and Granite, reflexive retrieval accounts for under 10\% of legible predictions, the lowest of any model we test. Nearly half of the predictions for these models fall outside the three mechanisms. Exploring whether this reflects a retrieval mechanism outside our taxonomy, model scale, or anything else, we leave for future work.

\end{document}